\documentclass[letterpaper, 10 pt, conference]{ieeeconf}  

\IEEEoverridecommandlockouts                              
\usepackage{graphics} % for pdf, bitmapped graphics files
\usepackage{epsfig} % for postscript graphics files
\usepackage{mathptmx} % assumes new font selection scheme installed
\usepackage{amsmath} % assumes amsmath package installed
\usepackage{amssymb}  % assumes amsmath package installed
\usepackage{url}
\usepackage{booktabs} 
\usepackage{graphicx} % for including images

\usepackage{xcolor}
\usepackage{float}
\usepackage{soul}
\title{\LARGE \bf
Phone2Act: A Low-Cost, Hardware-Agnostic Teleoperation System \\ for Scalable VLA Data Collection
}

\author{Om Mandhane$^{1*}$, Bipin Yadav$^{1*}$, Sangeetha Prasanna Ram$^{1}$, and Gopalakrishnan Narayanan$^{1}$%
\thanks{*These authors contributed equally to this work.}%
\thanks{This research was supported by Vivekanand Education Society's Institute of Technology (VESIT).}%
\thanks{$^{1}$All authors are with the Dept. of Automation \& Robotics Engineering, Vivekanand Education Society's Institute of Technology (VESIT), Mumbai, India. 
        {\tt\small \{ommandhane27, bipinyd27\}@gmail.com}, 
        {\tt\small \{sangeeta.prasannaram, n.gopalkrishnan\}@ves.ac.in}}%
}
\makeatletter
\let\@oldmaketitle\@maketitle
\renewcommand{\@maketitle}{\@oldmaketitle
  \vspace{-0.2cm} % Pulls the image slightly closer to the authors
  \begin{center}
    \includegraphics[width=0.85\textwidth]{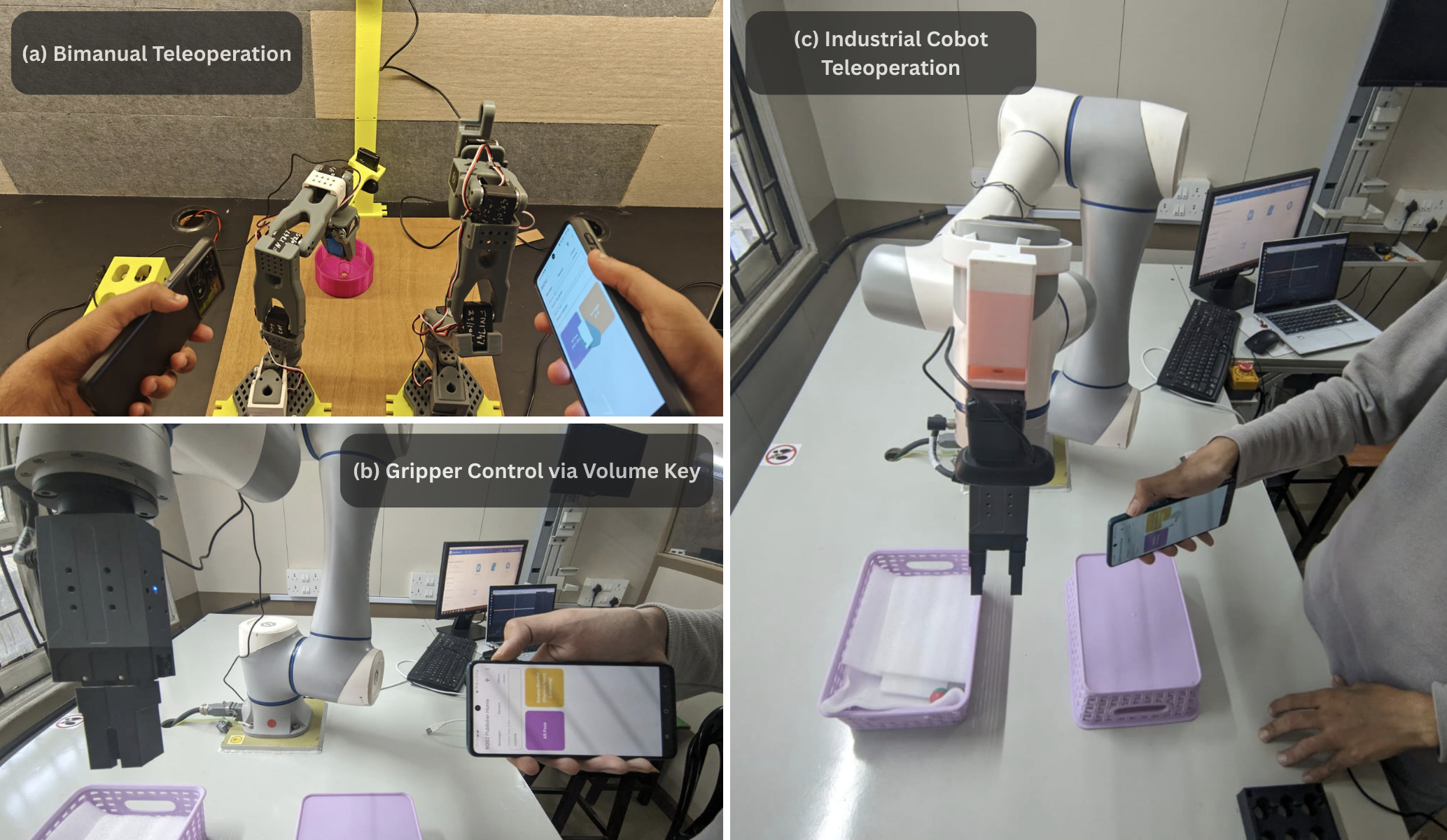} 
    \vspace{-0.1cm}
    \def\@captype{figure}
    \caption{\textbf{The Phone2Act Framework.} By transforming commodity smartphones into 6-DoF spatial controllers, the system provides a seamless, hardware-agnostic teleoperation interface. \textbf{(a)} The framework scales instantly to a low-cost, bimanual LeRobot SO-101 setup without requiring any core software modifications. \textbf{(b)} The custom Android interface utilizes hardware volume keys to actuate the gripper, allowing uninterrupted spatial control. \textbf{(c)} A user teleoperates an industrial Dobot CR5 for a high-precision pick-and-place task, demonstrating the system's viability across disparate hardware platforms.}
    \label{fig:teaser}
  \end{center}
  \vspace{0.05cm} % Adds a clean gap before the Abstract starts
}
\makeatother
\begin{document}

\maketitle
\thispagestyle{empty}
\pagestyle{empty}

\begin{abstract}
Collecting diverse, high-quality manipulation data for 
Vision-Language-Action (VLA) model training remains 
prohibitively expensive for many research groups, as 
existing teleoperation frameworks rely on specialized 
hardware or are tightly coupled to specific robot platforms. 
We present Phone2Act, a low-cost, hardware-agnostic 
teleoperation framework that transforms a commodity 
smartphone into a 6-DoF robot controller via Google ARCore. 
Built on a modular ROS 2 architecture, Phone2Act decouples 
control logic from hardware specifics through interchangeable 
bridge nodes, supporting platforms from industrial cobots to 
low-cost bimanual arms without code modification. A Universal 
Recorder synchronizes multi-camera RGB streams with robot 
state feedback and exports demonstrations natively in the 
LeRobot dataset format, eliminating post-processing and 
enabling immediate VLA fine-tuning. We validate the framework 
by fine-tuning GR00T-N1.5 on 130 collected episodes, 
achieving a 90\% success rate on a real-world multi-stage 
pick-and-place task deployed on a physical Dobot CR5.
\end{abstract}

\section{INTRODUCTION}
The paradigm shift towards generalist robot policies, exemplified 
by Vision-Language-Action (VLA) models such as Octo~\cite{octo2024} 
and OpenVLA~\cite{kim2024openvla}, has created an immense demand for 
diverse, high-quality manipulation datasets. Unlike domains where data 
can be scraped from the web, robotic data must be physically collected 
through teleoperation. However, for many research labs and educational 
institutions, this requirement presents a significant economic and 
technical bottleneck.

Current state-of-the-art data collection often utilizes bimanual 
``leader-follower'' systems such as Mobile ALOHA~\cite{fu2024mobilealoha}. 
While intuitive, these systems effectively double hardware costs by 
requiring a non-actuated replica arm for every active robot. Alternative 
interfaces such as VR headsets~\cite{meng2023virtual} offer immersive 
control but require expensive proprietary peripherals and complex spatial 
calibration. Furthermore, most existing teleoperation frameworks are 
hard-coded for specific commercial platforms, leaving researchers working 
with custom or non-standard hardware with no viable path to high-quality 
data collection without significant engineering overhead.

We present Phone2Act, a low-cost, hardware-agnostic teleoperation 
framework that transforms a commodity smartphone into a high-frequency 
6-DoF controller for robotic manipulation. Built on a modular ROS 2 
architecture, Phone2Act decouples control logic from hardware specifics 
through standardized topic interfaces, enabling integration across diverse 
manipulators — from industrial cobots to custom open-source arms — by 
simply swapping interchangeable bridge nodes. Crucially, the system 
includes a Universal Recorder that natively exports synchronized 
demonstrations directly in the LeRobot dataset format, eliminating 
post-processing and making collected data immediately ready for VLA 
fine-tuning.

The main contributions of this work are:
\begin{itemize}
    \item A modular, hardware-agnostic ROS 2 teleoperation framework 
    controlled via commodity smartphone, requiring no specialized 
    peripherals or per-platform re-engineering.
    \item A Universal Recorder that natively exports synchronized 
    multi-camera demonstrations in the LeRobot dataset format, 
    ready for VLA fine-tuning without post-processing.
    \item Experimental validation across single-arm and bimanual 
    platforms, culminating in a 90\% real-world success rate on a 
    pick-and-place task using the GR00T N1.5 VLA model~\cite{bjorck2025groot} trained 
    on 130 Phone2Act episodes.
\end{itemize}
To promote reproducibility and community adoption, the 
Phone2Act Android application, ROS 2 teleoperation framework, and example datasets will be released under the MIT License upon publication.
The remainder of this paper is structured as follows: Section II 
reviews related work; Section III details the system 
architecture; and Section IV presents experimental 
results and validation.

\section{RELATED WORK}

Smartphone-based teleoperation has emerged as a low-cost, accessible interface for robotic manipulation, with prior systems spanning inertial control, crowdsourced demonstration collection, and real-world mobile manipulation. However, none of these frameworks were designed around a robot-agnostic dataset abstraction suitable for scalable Vision-Language-Action (VLA) model training.

Early work established the viability of smartphones as control interfaces. Parga et al.~\cite{parga2013tele} demonstrated that accelerometer and gyroscope readings from a commodity device could drive a robotic arm, showing that inertial sensing alone was sufficient for basic teleoperation. Similarly, Wu et al.~\cite{wu2020development} explored smartphone interfaces for accessibility applications, introducing reduced-degree-of-freedom control schemes to lower the barrier for non-expert operators. These systems treated the smartphone as a human--machine interface, with no consideration for structured data logging or learning pipelines.
\begin{figure}[t] % [t] pushes it to the top of the column
    \centering
    \includegraphics[width=0.6\linewidth]{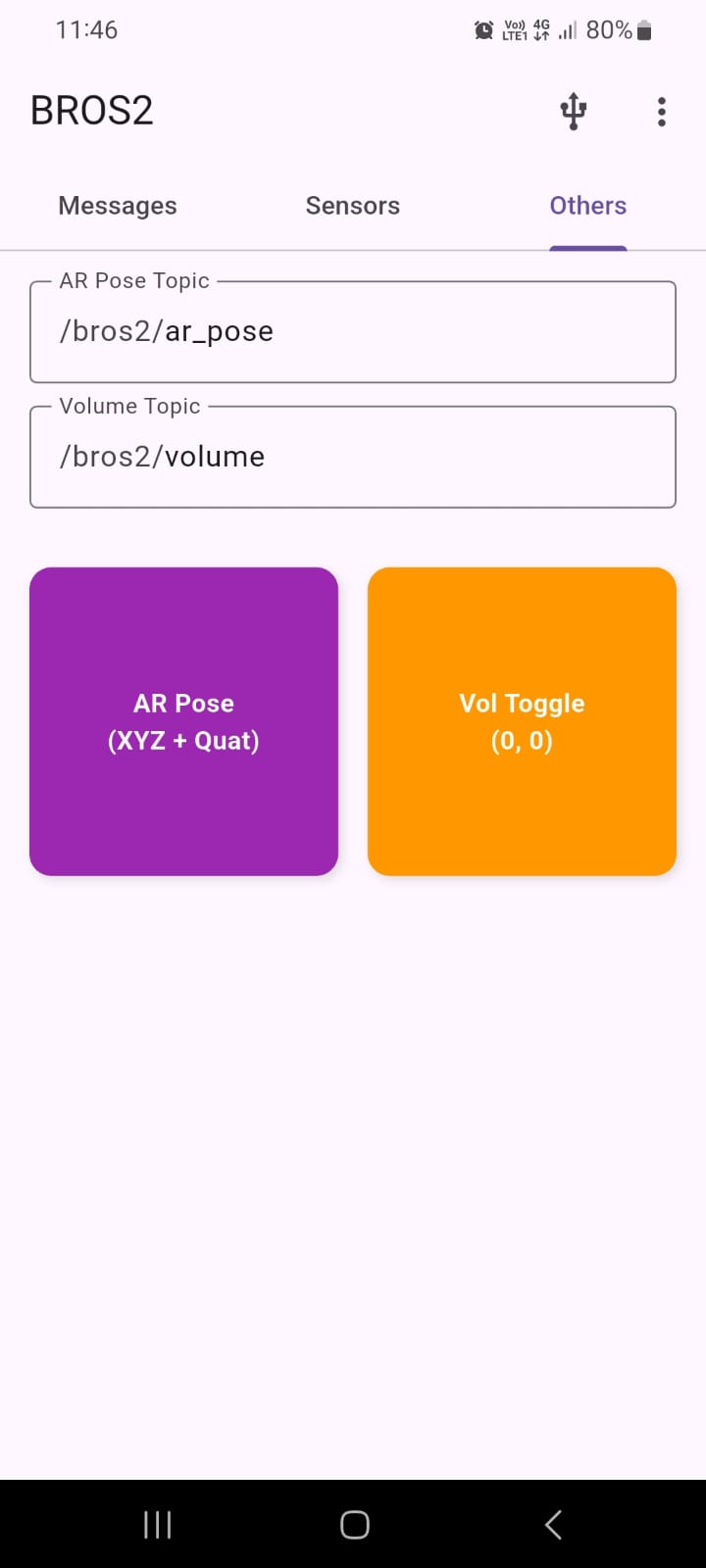} 
    \caption{\textbf{Mobile Interface.} The custom Phone2Act Android application featuring configurable ROS 2 topics.}
    \label{fig:ui}
\end{figure}
The focus shifted toward scalable demonstration collection with RoboTurk~\cite{mandlekar2018roboturk}, which enabled crowdsourced 6-DoF trajectory recording using ARKit-based visual--inertial pose tracking and WebRTC streaming. MART~\cite{tung2021learning} extended this paradigm to collaborative multi-arm settings, enabling study of centralized and decentralized policy architectures, while MoMaRT~\cite{wong2021error} further generalized it to mobile manipulation tasks. Concurrently, Werner and Melek~\cite{werner2024fiveg} evaluated 5G-networked smartphone teleoperation, focusing on latency and telepresence quality. Most recently, TidyBot++~\cite{wu2024tidybotpp} paired smartphone teleoperation with an open-source holonomic mobile manipulator and demonstrated successful diffusion-policy training from real-world demonstrations.

Collectively, these systems demonstrate significant progress in teleoperation capability and scale. Yet each is tightly coupled to its own robot platform, simulation environment, or training pipeline. None provides a standardized, embodiment-agnostic logging format suitable for cross-platform dataset aggregation or VLA fine-tuning.

Phone2Act addresses this gap through a data-first design. Rather than centering on teleoperation hardware or policy architecture, it introduces a Universal Recorder that synchronizes external RGB camera streams with robot state feedback at runtime and directly exports episodes in the standardized LeRobot format (Parquet and MP4). By abstracting embodiment-specific interfaces through ROS 2, Phone2Act enables a single teleoperation pipeline to generate training-ready datasets across diverse manipulators---from industrial arms to low-cost bimanual setups.

\section{SYSTEM ARCHITECTURE}

\begin{figure*}[t]
  \centering
  \includegraphics[width=0.9\textwidth]{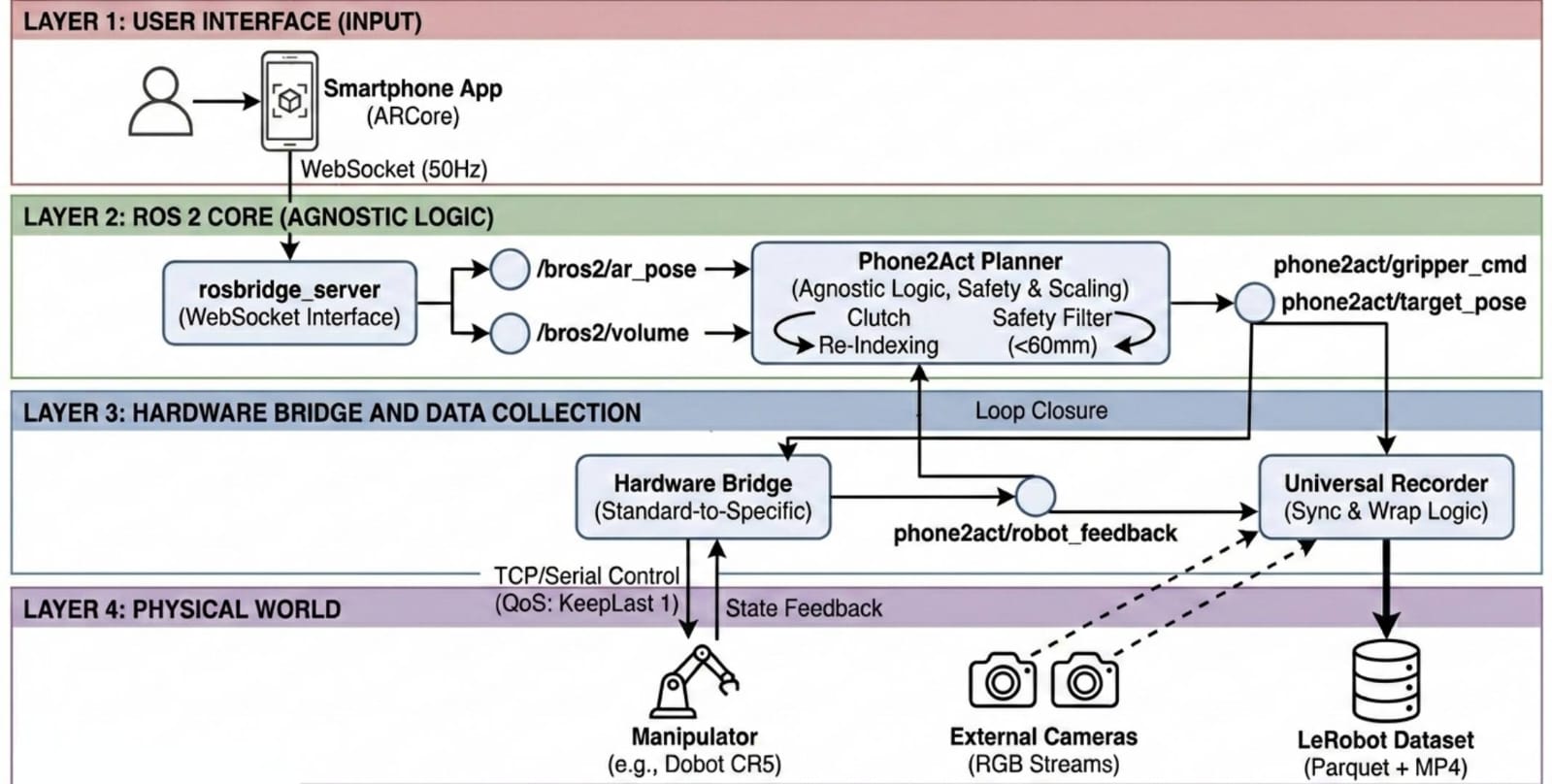}
  \caption{Phone2Act Architecture. The system decouples the 
smartphone input from the robot hardware via a central, 
agnostic planner and standardized ROS 2 topics 
(\texttt{phone2act/target\_pose}, 
\texttt{phone2act/gripper\_cmd}, and 
\texttt{phone2act/robot\_feedback}).}
  \label{fig:arch}
\end{figure*}
Phone2Act is built on a modular ROS 2 framework organized into three layers, as illustrated in Fig.~\ref{fig:arch}: (1) an \textit{Interface Layer} that captures smartphone pose and button events, (2) an \textit{Agnostic Core} that transforms raw phone input into robot-independent motion commands, and (3) a \textit{Hardware Bridge and Data Collection Layer} that bridges those commands to a specific robot and records synchronized demonstration data. Each layer communicates exclusively through standardized ROS 2 topics, so any component can be swapped without modifying the others.

\subsection{The Interface Layer (Smartphone)}

A custom Android application leverages Google's ARCore to provide stable 6-DoF pose estimation relative to its starting position. The app publishes this pose data, along with volume button events, at 50Hz to a \texttt{rosbridge\_server} via WebSocket. The volume keys serve as intuitive binary inputs: Volume Up 
engages the clutch for repositioning, and Volume Down toggles 
the gripper, publishing a binary command to 
\texttt{phone2act/gripper\_cmd}.

\subsection{The Agnostic Core (Planner)} 
The \texttt{phone2act\_planner} node sits at the center of the 
pipeline. It subscribes to raw phone pose and button topics and 
publishes standardized absolute Cartesian target poses to 
\texttt{phone2act/target\_pose}, knowing nothing about which 
robot is downstream. Four mechanisms inside the planner make this mapping robust:

\begin{itemize}
    \item \textbf{Spatial Mapping and Frame Alignment:}
    To translate local phone movements into the robot's base frame, the planner applies a static axis-alignment transformation. Following initialization or a clutch release, the system continuously calculates the translation delta $\Delta P = [\Delta x_p, \Delta y_p, \Delta z_p]^T$ relative to the phone's reference origin. This delta is mapped to the robot's Cartesian target $R_{target}$ via a static axis-alignment matrix $M$, such that $R_{target} = R_{initial} + M(\Delta P \cdot S)$, where $S$ is a configurable scaling factor. For our configuration, $M$ aligns the phone's landscape orientation with the robot's workspace (e.g., mapping phone-forward to robot-base-$X$), ensuring intuitive directional parity for the operator across disparate hardware.
    \item \textbf{Floating Zero (Clutching):}
    A smartphone has a smaller physical workspace than a robot arm, and operators naturally need to pause or reposition the device mid-task. Both needs are addressed by a single toggle-based re-indexing mechanism. The first button press \textit{engages} the clutch: command output is suspended and the robot holds its position, freeing the operator to move the phone arbitrarily — to rest, reposition, or hand the device to another person. The second press \textit{releases} the clutch: the current phone pose and robot end-effector pose are jointly recorded as a new reference origin, and teleoperation resumes from that point, giving the operator an effectively unlimited virtual workspace.

    \item \textbf{Safety and Workspace Enforcement:} 
    To protect the robot from physical damage and erratic behavior, the planner implements a dual-stage validation filter. First, a \textbf{Workspace Clamp} restricts all incoming commands to a \textit{user-configurable} safe Cartesian volume $[X_{min,max}, Y_{min,max}, Z_{min,max}]$, ensuring the system can be instantly tuned to any manipulator's mechanical limits to prevent self-collisions or base impacts. Second, a \textbf{Zero-Jump Filter} handles input anomalies: tracking loss, an accidental device drop, or transient network glitches can all produce large discontinuous jumps in the incoming pose stream. The planner monitors the Euclidean distance between consecutive target positions; any jump exceeding a configurable threshold (e.g., 60\,mm) is silently dropped. Recovery is automatic—normal operation resumes as soon as the device returns within bounds—ensuring that a momentary drop or glitch does not require a system restart.
    
    \item \textbf{Euler-Space Rotation Composition:} 
    Poses are transported as quaternions over standard \texttt{PoseStamped} messages, but rotation \textit{commands} are computed internally as roll-pitch-yaw (RPY) deltas. Unlike quaternions, RPY imposes no unit-norm constraint, making it better suited for direct neural network regression. This representation is deliberate: by computing and storing rotation as RPY deltas, the teleoperation action space matches exactly the continuous RPY space over which downstream VLA policy heads regress at inference time~\cite{octo2024, kim2024openvla, wen2025tinyvla, black2024pi0}. At 50\,Hz with small per-step deltas, gimbal-lock singularities are avoided in practice. The planner converts incoming quaternions to RPY, accumulates the delta, then converts back to quaternion solely for transport.
\end{itemize}

\subsection{Hardware Bridge \& Data Collection}
%---------------------------------------------------------------
This layer contains two independent nodes. The \textit{Hardware Bridge} closes the control loop with the physical robot, while the \textit{Universal Recorder} — the primary data contribution of this work — captures synchronized demonstration episodes directly in a standard VLA-ready format.

\subsubsection{Hardware Bridge}
Hardware specifics are encapsulated in interchangeable bridge nodes. For the Dobot CR5 used in our experiments, the \texttt{dobot\_bridge} node handles three responsibilities.

\textbf{Unit and Format Translation.} The bridge converts the planner's meter-scale quaternion commands into the millimeter/Euler-degree format expected by the Dobot CR5 real-time port, sending commands over a TCP socket with \texttt{TCP\_NODELAY} to minimize latency.

\textbf{Command Freshness.} Control topics are subscribed with a QoS profile of \texttt{BEST\_EFFORT} reliability and queue depth of 1 (\texttt{KEEP\_LAST}), ensuring stale commands are discarded during network congestion so the robot always acts on the operator's most recent intent.

\textbf{State Feedback.} The bridge continuously reads the robot's actual end-effector pose, normalizes it back to meters and quaternions, and publishes it to \texttt{phone2act/robot\_feedback}, closing the loop for both the planner's re-indexing logic and the recorder's ground-truth state capture.

To support rapid integration of new platforms, the framework ships a generic bridge template that pre-wires the required QoS profiles and topic interfaces; researchers need only insert their robot's API calls for communication and kinematics.

\subsubsection{Data Collection Pipeline (Universal Recorder)}
The \texttt{universal\_recorder} node listens exclusively to standard ROS 2 topics, making it entirely hardware-agnostic. It captures synchronized RGB frames from external webcams alongside robot joint states and end-effector poses at 20\,Hz. A strict temporal synchronization gate ensures a frame is recorded only when fresh data is simultaneously available from all sources, preventing incomplete or misaligned timesteps.

A known challenge in Euler-angle action spaces is the discontinuity at $\pm180^\circ$, where a minimal physical rotation near a boundary registers as a large numerical jump, causing gradient instabilities during training. The recorder mitigates this by applying shortest-path wrapping to rotational deltas:
\begin{equation}
\Delta\theta_{\mathrm{wrapped}} = (\Delta\theta + \pi) \bmod 2\pi - \pi
\end{equation}
This guarantees smooth, bounded rotational targets across the full action space.

At each timestep, the recorder constructs a per-timestep observation vector comprising proprioceptive joint positions, Cartesian end-effector pose, and gripper state, alongside a corresponding action vector of 6-DoF Cartesian deltas and target gripper state. To avoid I/O overhead during the time-critical recording loop, video frames are buffered entirely in RAM and flushed to MP4 only upon episode completion, preserving the 20\,Hz timing guarantee. Processed episodes are then serialized directly into the LeRobot dataset format—visual observations as MP4 videos and episode metadata as Parquet files—eliminating any post-processing step and allowing datasets to be immediately ingested by standard VLA training pipelines.

\section{EXPERIMENTS \& RESULTS}

We validated the Phone2Act system through latency analysis, bimanual deployment, and a real-world VLA training task.

\subsection{System Latency}
We measured the true end-to-end latency—from physical smartphone translation to the onset of observable robotic actuation—using high-speed video analysis at 240 FPS. Across multiple spatial axes under standard 2.4 GHz Wi-Fi conditions, the integrated system exhibited an end-to-end latency ranging from approximately 350ms to 440ms (averaging ~395ms). 

It is critical to note that this measurement is heavily dominated by the hardware-level constraints of the test platform. Industrial collaborative arms like the Dobot CR5 intrinsically introduce significant delay due to mechanical inertia and internal trajectory smoothing for safety. Consequently, the actual software latency contributed by the Phone2Act pipeline (ARCore tracking, WebSocket transmission, and ROS 2 processing) represents only a small fraction of this total. This demonstrates that our hardware-agnostic framework is highly optimized and introduces minimal overhead. Furthermore, while higher than hardwired direct-drive systems, this overall delay proved fully acceptable for quasi-static VLA data collection, as validated by our successful model fine-tuning and real-world deployment.

\subsection{Bimanual and Cross-Hardware Deployment}
To demonstrate hardware agnosticism and scalability, we deployed the system to control a pair of low-cost, 3D-printed LeRobot SO-101 arms. Crucially, scaling from a single-arm setup to a dual-arm configuration required zero modifications to the core source code. 

To prevent data interference between the two controllers, the framework leverages standard ROS 2 namespaces. A standard launch file spins up two identical instances of the planner and bridge nodes, placing them in logically isolated namespaces (\texttt{/left} and \texttt{/right}). Concurrently, a dynamically configurable topic setting within the Android application allows the user to assign each smartphone to a specific arm at runtime. By instructing one device to publish to the \texttt{/left} namespace and the other to the \texttt{/right} namespace, the dual data streams are kept perfectly distinct. This setup successfully achieves synchronized bimanual teleoperation, replicating the capability of a high-end dual-arm station at a fraction of the cost.

\subsection{VLA Policy Learning (GR00T)}
We collected a dataset of 130 teleoperation episodes using a Dobot CR5 for a multi-stage manipulation task: "Pick up the ball and place it in the purple basket/tray." The intuitive smartphone interface enabled high-speed data acquisition, averaging 2–3 episodes per minute without causing operator fatigue. To validate the quality and compatibility of the Phone2Act data, we then fine-tuned the \textbf{NVIDIA GR00T-N1.5-3B} Vision-Language-Action (VLA) foundation model.

The fine-tuning process was executed on a single NVIDIA RTX A6000 GPU using the native training pipeline. To optimize compute efficiency and align the policy with the specific robotic embodiment, we employed a selective parameter-tuning strategy: the core Large Language Model (LLM) backbone was kept completely frozen, while gradients were applied exclusively to the visual encoder and the diffusion-based action head. 
We utilized an effective batch size of 48 and a peak learning rate of $1 \times 10^{-4}$ with bfloat16 mixed precision. The model was configured to predict action chunks with a horizon of 10 steps across a 7-dimensional action space—consisting of a 6-DoF end-effector pose and a binary gripper command—enabling smooth and continuous trajectory generation. The model rapidly internalized the task logic, demonstrating clear convergence as the diffusion training loss (MSE) plateaued near $0.05$. Consequently, training was successfully concluded at 2,000 steps, and this converged policy was utilized for all subsequent evaluation and deployment.

\begin{figure}[thpb]
  \centering
  \includegraphics[width=\linewidth, height=0.85\textheight, keepaspectratio]{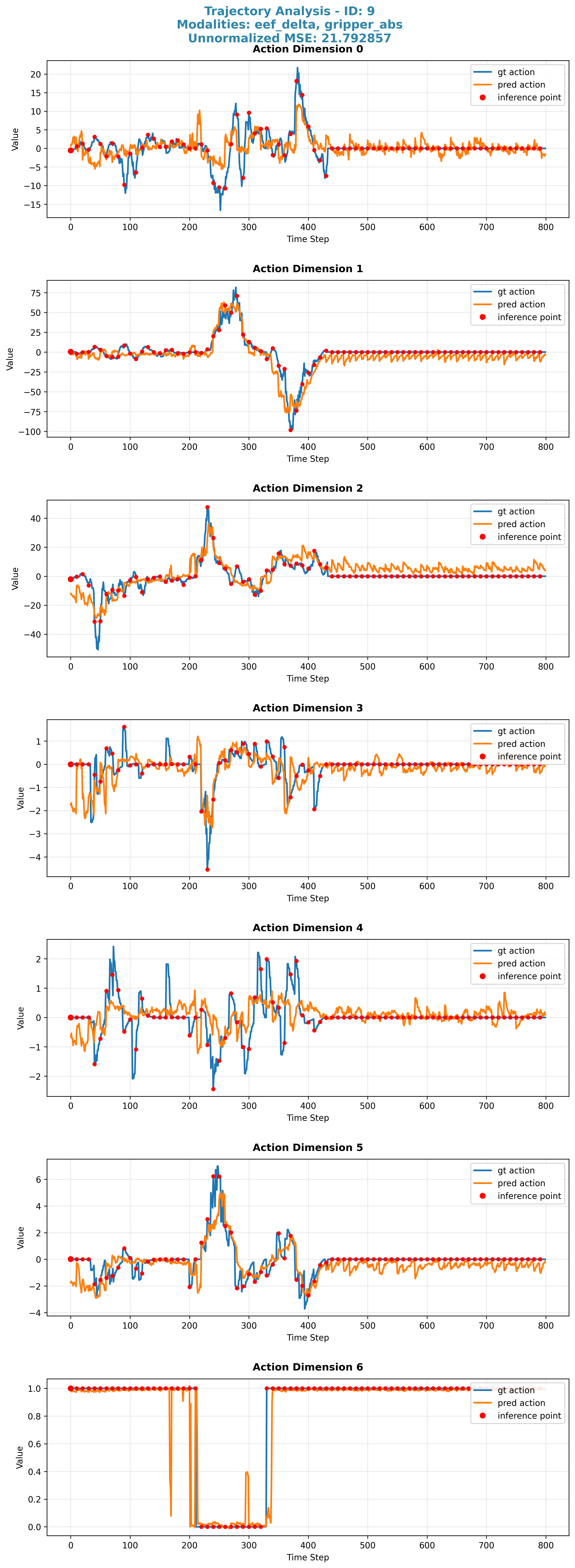}
  \caption{Open-Loop Evaluation of Groot N1.5 trained on Phone2Act data. The predicted actions (orange) closely track the ground truth teleoperation data (blue), indicating successful learning of control signals.}
  \label{fig:mse}
\end{figure}

\subsubsection{Open-Loop Trajectory Analysis}
To verify the model's ability to internalize the teleoperated demonstrations, we conducted an open-loop evaluation on held-out validation trajectories. As shown in Fig. \ref{fig:mse}, the predicted actions (orange) successfully capture the temporal dynamics and shape of the human ground-truth data (blue) across the complete action space (6-DoF Cartesian deltas and binary gripper actuation). While standard open-loop distance metrics like unnormalized Mean Squared Error often fail to reflect true closed-loop performance in imitation learning due to compounding covariate shift, the strong qualitative alignment observed here—particularly in predicting critical inflection points and gripper state transitions—served as a robust indicator of policy convergence prior to hardware deployment.

\begin{figure}[t]
  \centering
  \includegraphics[width=\linewidth]{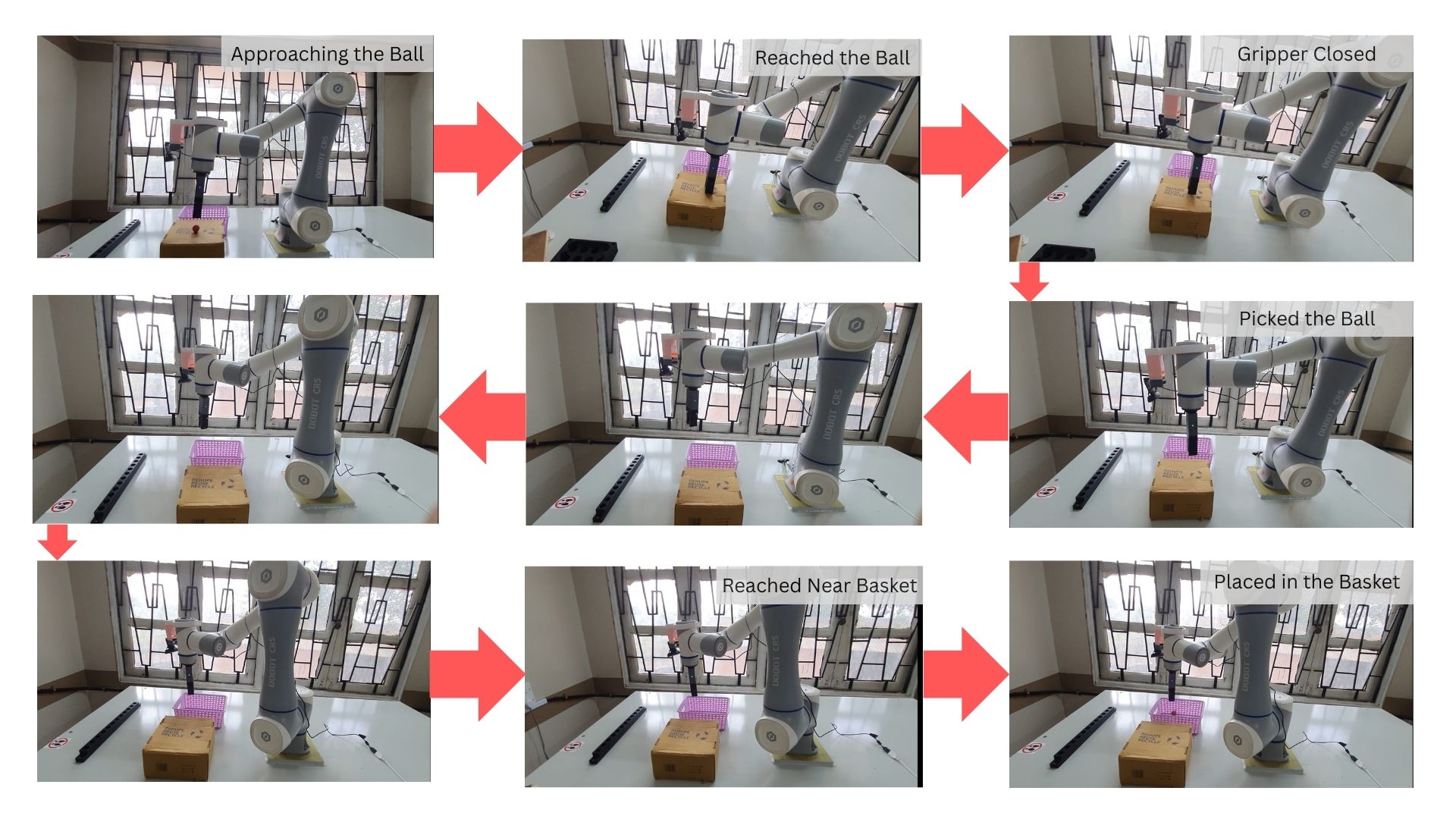}
  \caption{Real-world execution of the learned policy. \textbf{Task:} \textit{"Pick up the ball and place it in the purple basket."} The Dobot CR5 smoothly executes the sequence: approaching, grasping the target, and placing it into the designated basket.}
  \label{fig:real_world}
\end{figure}
\subsubsection{Real-World Deployment}
The ultimate validation of the Phone2Act framework was the deployment of the fine-tuned GR00T-N1.5-3B policy on the physical Dobot CR5. To accommodate the heavy compute requirements of the 3-billion parameter model, inference was handled via a cloud-based client-server node, seamlessly integrating with our existing ROS 2 hardware bridge without requiring additional low-level tuning. 

The model was evaluated on the multi-stage task of picking up a target ball and placing it into a designated basket. Across 10 consecutive real-world trials, the system achieved 9 successful executions (90\% success rate). A representative successful episode is visualized in Fig. \ref{fig:real_world}. The policy demonstrated robust closed-loop behavior, smoothly correcting for minor perturbations. This high success rate confirms that the data collected via the 50Hz Phone2Act pipeline provides sufficient fidelity to train capable, real-world VLA policies.

%%%%%%%%%%%%%%%%%%%%%%%%%%%%%%%%%%%%%%%%%%%%%%%%%%%%%%%%%%%%%%%%%%%%%%%%%%%%%%%%
\section{CONCLUSION}

Phone2Act provides a scalable, low-cost solution for the data-intensive requirements of VLA robotics. By decoupling the input interface from hardware specifics through a standardized ROS 2 architecture, the framework ensures easy integration across diverse manipulators while providing a direct pipeline to record high-quality data for VLA training in the LeRobot format. This modularity significantly lowers the barrier to entry for collecting large-scale, high-fidelity manipulation datasets. Future work will focus on integrating haptic feedback via the smartphone’s vibration motor to further enhance operator situational awareness.

\section*{ACKNOWLEDGMENT}
This research was supported by Vivekanand Education So-
ciety’s Institute of Technology (VESIT). The authors would
like to thank Dr. Debanik Roy for their valuable feedback
and review of the manuscript.
The system architecture diagram (Fig.~\ref{fig:arch}) was designed by the authors and rendered into its final visual form with the assistance of Gemini. The authors acknowledge the use of Google Gemini for assistance with copy-editing, academic phrasing refinement, and LaTeX formatting in the Introduction, Related Work, and System Architecture sections. All technical content, system design, experiments, and conclusions are the sole work of 
the authors.

% \addtolength{\textheight}{-12cm}  % Balance columns on last page

\typeout{get arXiv to do 4 passes: Label(s) may have changed. Rerun}
\end{document}